\documentclass[letterpaper, 10 pt, conference]{ieeeconf}
\IEEEoverridecommandlockouts
\usepackage{cite}
\usepackage{amsmath,amssymb,amsfonts}
\usepackage{algorithmic}
\usepackage{graphicx}
\usepackage{booktabs}
\usepackage{multirow}
\usepackage{textcomp}
\usepackage{xcolor}
\usepackage{url}

\newcommand{\std}[1]{{\tiny$\pm$#1}}

\begin{document}

\title{Think Like a World Model, Act Like a VLA:\\
Distilling World-Model Representations into Compact Robot Policies
}

\author{Trung Dao$^{1}$, Sankalp Yamsani$^{2}$, Jaden Park$^{1}$, Joohyung Kim$^{2}$, and Yong Jae Lee$^{1}$%
\thanks{$^{1}$University of Wisconsin--Madison, USA.
        {\tt\small tdao6@wisc.edu, \{jadenpark, yongjaelee\}@cs.wisc.edu}}%
\thanks{$^{2}$University of Illinois Urbana-Champaign, USA.
        {\tt\small \{yamsani2, joohyung\}@illinois.edu}}%
}

\maketitle

\begin{abstract}
Vision-Language-Action (VLA) models map observations to actions with no objective that accounts for how the world responds, so their robustness is bounded primarily by data coverage. World models carry precisely that missing objective and are better grounded for it, yet rolling the future forward costs seconds per decision and rules them out of the control loop. We show the two can be separated. What a world model knows about physical scenes lives in its \emph{internal features}; generating the future is merely the objective that produced them, so the grounding can be inherited while the generative machinery is left behind. We add one feature-alignment term to ordinary VLA training: a frozen world model is run over the training frames once and cached, and the student learns to agree with that cache. No teacher is loaded during training, the projector is discarded after it, and the deployed policy is identical to the undistilled baseline, running in $32$~ms and $1.86$~GB on a consumer RTX~5090, so every gain is attributable to the representation rather than to added capacity or test-time compute. A $0.8$B student reaches $97.9\%$ on LIBERO, improves from $48.2\%$ to $50.5\%$ on RoboCasa-GR1 humanoid manipulation, and the same objective carries over to real hardware, on both a single-arm and a bimanual platform. The gain survives changes of student scale, backbone, alignment layer, and teacher, indicating a broad representational prior rather than a fragile alignment between two particular networks. Project page: \url{https://thaw-vla.trung-dt.com/}.
\end{abstract}


\section{Introduction}
\label{sec:intro}

Vision-Language-Action (VLA) models are a compelling paradigm for general-purpose robot control: by using large vision-language backbones as the foundation and training them to predict robot actions, they inherit strong semantic understanding and scale efficiently to diverse tasks \cite{kim2025openvla,black2025pi0,black2025pi05,gr00t}. In practice, however, VLAs are brittle to environmental perturbations such as viewpoint shifts, lighting changes, and clutter that are underrepresented in their training data \cite{zhang2026wam}, and adapting them to unseen tasks or new embodiments takes a large amount of additional data \cite{dreamzero}. The cause is structural. Standard VLA training maps observations to actions with no explicit pressure to model how the world evolves in response to those actions, so a policy can fit its training distribution well without ever learning temporally grounded or causally consistent representations. Robustness then becomes primarily a function of data coverage, which is expensive and ultimately insufficient.

\begin{figure}[!t]
    \centering
    \includegraphics[width=\columnwidth]{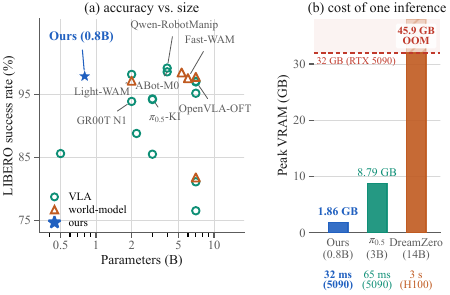}
    \caption{\textbf{A $0.8$B policy at the accuracy of models several times its size.} \textbf{(a)} LIBERO four-suite mean success rate against parameter count. \textbf{(b)} What one inference costs, against the $32$~GB of a consumer RTX~5090.}
    \label{fig:teaser}
\end{figure}

World models take the opposite approach. Instead of learning only to emit controls, they are trained to model the future environment, often by jointly predicting future frames, states, and actions. This additional burden forces the model to internalize temporal structure and causal consequence, yielding representations that are more physically grounded than those of an action-only policy. Recent work reports that policies built this way transfer better to new tasks \cite{dreamzero,kim2026cosmos,li2026lingbot} and degrade far more gracefully under perturbation \cite{zhang2026wam}. The benefit comes at a practical cost: rolling the world forward is much more expensive than predicting an action directly. DreamZero \cite{dreamzero} requires 3s per inference on an H100, given several layers of optimization (caching, kernel fusion, quantization, distillation) to become usable, whereas $\pi_{0.5}$ \cite{black2025pi05} runs in 65ms on a consumer RTX~5090 and our own $0.8$B policy in 32ms on the same card (Fig.~\ref{fig:teaser}), at 1.86GB against DreamZero's 45.9GB, which the card cannot hold. A world model is therefore attractive as a source of knowledge and (currently) unattractive as a deployed policy.

This tension motivates our question: \textit{can a lightweight VLA be trained to inherit the grounded representations of a large world model without paying its inference cost?} We answer yes, and with a mechanism far cheaper than the question suggests. The key observation is that what a world model knows about physical scenes is already present in its \emph{internal features}; the expensive part, generating the future, is only the objective that produced them. A student can therefore be supervised on the features alone.

We instantiate this as a representation-alignment objective on the student's visual pathway, with a frozen world-model supplying the target. Two design choices make the objective cheap. Because the teacher is frozen, its targets depend only on the training frames and can be extracted once, ahead of time, and read back during training instead of recomputed; and because the student is asked to agree with those targets in \emph{direction} rather than reproduce them exactly, it is free to retain whatever additional structure its action objective demands, and the two backbones need not share a feature space or a dimensionality. What follows is a distillation procedure that is invisible from both ends. Training never holds a teacher in memory, so a distilled run costs what the baseline run costs; the alignment head is auxiliary and is discarded when training ends; and the deployed policy is architecturally identical to the undistilled baseline, down to the number of flow steps. The grounding is inherited during training and carried for free thereafter. Sec.~\ref{sec:method} gives the details.

The payoff is that a small policy behaves like a much larger one. Our $0.8$B student improves from $48.2\%$ to $50.5\%$ on RoboCasa-GR1 humanoid manipulation, within $4.3$ points of the $4$B model of the same architecture, matches a $4$B $\pi$-style policy at $28/30$ on a real single-arm pick-and-place task while carrying a fifth of its parameters, and reaches $97.9\%$ on LIBERO, $2.6$ points above the identical undistilled student. Our contributions are:

\begin{itemize}
    \item \textbf{A distillation recipe that transfers world-model grounding at
    zero inference cost.} One cosine term against a cached teacher feature, no teacher forward pass during training, and no change to the deployed graph (Sec.~\ref{sec:method}).

    \item \textbf{A $0.8$B policy competitive with much larger ones.}
    Consistent gains on a humanoid simulation benchmark and on real
    single-arm and bimanual hardware, with the undistilled policy of
    identical size as the control
    (Secs.~\ref{sec:exp_gr1} and~\ref{sec:exp_real}).

    \item \textbf{Ablations showing the recipe is robust to its own design
    choices.} The gain survives changes of student scale, student backbone,
    alignment layer, and teacher family, which indicates the transferred
    signal is a broad representational prior rather than a delicate alignment
    between two particular networks (Sec.~\ref{sec:exp_ablation}).
\end{itemize}

\section{Related Work}
\label{sec:related}

\subsection{Vision-Language-Action Models}
VLA models are motivated by the observation that large vision-language models pretrained on internet-scale data already contain rich perceptual and semantic representations useful for robot manipulation. By adapting these models to predict control actions, recent work has shown that visuomotor policies can inherit broad semantic understanding and generalize across diverse tasks.

OpenVLA \cite{kim2025openvla} provides one of the earliest open-source works that validated this paradigm, using a 7B Llama-2 backbone. It discretizes continuous robot actions into bins, treats them as action tokens, and trains the model to autoregressively predict actions conditioned on visual observations and language instructions. More recent systems have extended this idea by adopting architectures designed specifically for efficient robot control. GR00T~N1 \cite{gr00t}, for example, adopts a dual-system architecture in which a large vision-language backbone produces a latent action plan that conditions a lightweight diffusion action expert, enabling dexterous bimanual manipulation across multiple tasks. Similarly, $\pi_0$ and $\pi_{0.5}$ \cite{black2025pi0,black2025pi05} build on PaliGemma \cite{paligemma} and adopt a comparable two-level design within a unified flow-matching framework: instead of separating the model into distinct modules, they integrate the action expert with the vision-language backbone by concatenating their query, key, and value representations and performing cross-attention at each layer. $\pi_{0.5}$ further improves generalization through co-training objectives that extend supervision beyond action prediction.

Our student comes from StarVLA \cite{starvla2026}, a modular codebase that pairs a common vision-language backbone with interchangeable action heads, giving a family of policies (QwenGR00T, QwenPI, QwenOFT, QwenFAST) that differ only in how actions are produced.

A parallel line of work pushes VLAs toward smaller budgets, which is the regime this paper targets. SmolVLA \cite{shukor2025smolvla} trains a 2.2B policy that runs on consumer hardware, VLA-OS \cite{gao2025vlaos} dissects planning representations at 0.5B, and Seer \cite{tian2025seer} couples predictive inverse dynamics with visual forecasting at 0.57B. These models trade accuracy for deployability; our aim is to recover the accuracy without giving up the budget.

\subsection{World Models}
World models are trained to predict how a scene evolves, without necessarily emitting actions. Video generative models are the dominant instance: Wan \cite{wan2025} is a large-scale video diffusion transformer, and DreamDojo \cite{gao2026dreamdojo0} learns a generalist robot world model from human video. Cosmos~3 \cite{cosmos3} is an omnimodal mixture-of-transformers that combines an understanding tower, a generative tower, and action and audio towers in one checkpoint. A related family learns predictive representations without pixel reconstruction: V-JEPA~2 \cite{assran2025vjepa2} pairs a self-supervised video encoder with an action-conditioned predictor trained in latent space, and serves as one of our ablation teachers.

\subsection{World Action Models}
World action models (WAMs) turn a world model into a policy, so that predicting the future and choosing an action are trained together. Cosmos-Policy \cite{kim2026cosmos} adapts a video-generation foundation model for control with minor architectural changes, folding robot state and value information into a shared latent that supports both action generation and future-state reasoning. WorldVLA \cite{cen2025worldvla} unifies action and image generation autoregressively in a single transformer. LingBot-VA \cite{li2026lingbot} and DreamZero \cite{dreamzero} instead decompose the problem into two stages, first predicting future frames and then using them as conditioning context for action prediction. Fast-WAM \cite{fastwam2026}, our second ablation teacher, builds a WAM on the Wan2.2-TI2V-5B \cite{wan2025} video diffusion transformer and is representative of the video-DiT branch of this family. More recent entries target the cost of the paradigm directly: OA-WAM \cite{oawam} decomposes frames into object-addressable slots, and Light-WAM \cite{lightwam} performs future-video supervision in a downsampled latent space with a compact backbone.


\subsection{Representation-Level Distillation}
Classical knowledge distillation transfers a teacher's output distribution \cite{hinton2015distilling}; FitNets \cite{romero2015fitnets} showed that intermediate ``hints'' can transfer more than outputs alone. REPA \cite{yu2025repa} revived this idea in generative modeling, showing that aligning an intermediate diffusion-transformer feature to a frozen self-supervised encoder dramatically accelerates training, using a purely point-wise, projection-mediated cosine objective with no generative component. Our alignment term performs representation-level distillation with a world model. The difference that matters here is what the frozen encoder knows: because its training objective demanded predicting how scenes evolve, its features carry temporal and causal structure, and the student inherits that structure without ever predicting a future itself.

\section{Method}
\label{sec:method}

\begin{figure*}[!t]
    \centering
    \includegraphics[width=\textwidth]{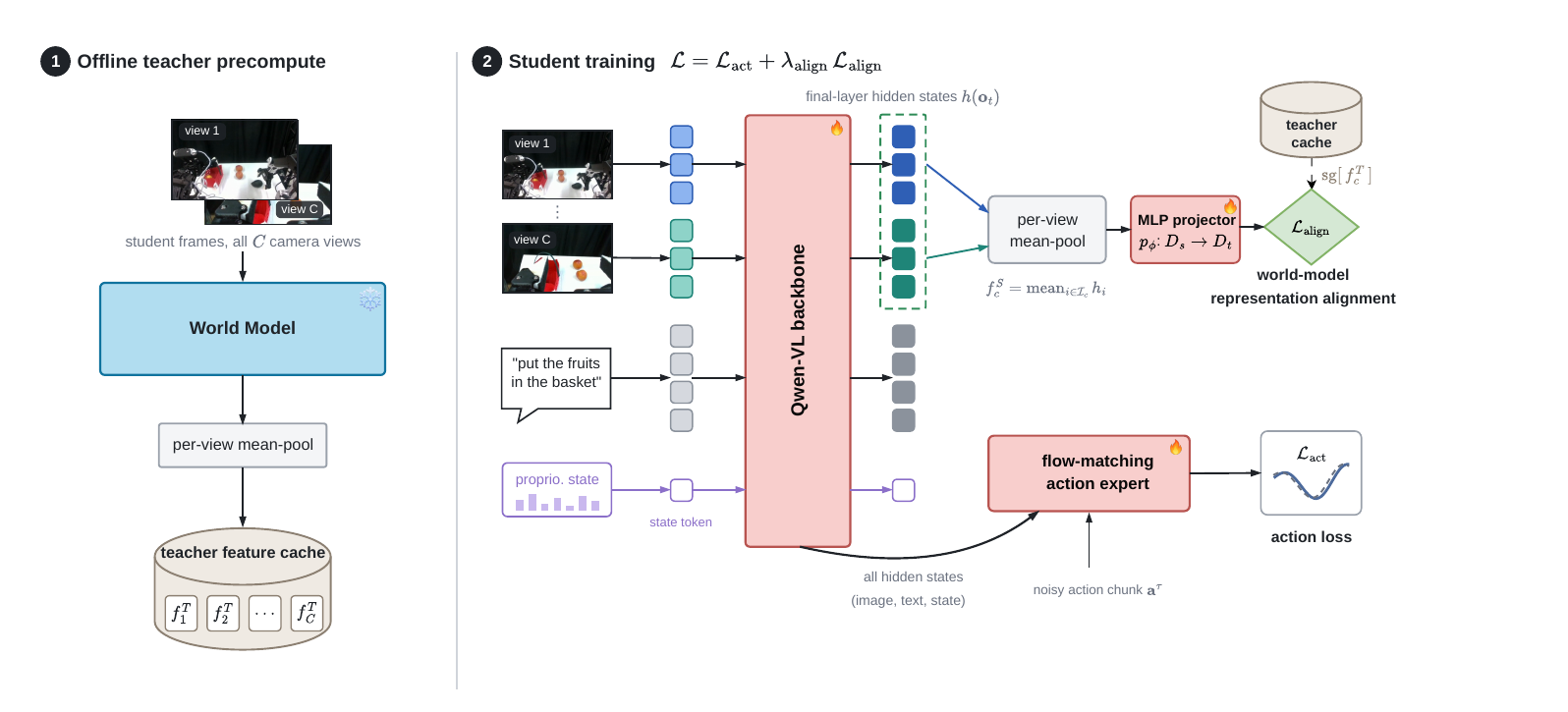}
    \caption{Overview of the method. \textbf{Left:} the frozen teacher world model feature extraction pipeline. \textbf{Right:} the training pipeline for the student VLA.}
    \label{fig:system_fig}
\end{figure*}

Our method is one term added to ordinary VLA training: align the student's pooled image-token features to those of a frozen world model (Fig.~\ref{fig:system_fig}). Sec.~\ref{sec:method_student} fixes the student and its usual action objective, Sec.~\ref{sec:method_c1} specifies the alignment term, and Sec.~\ref{sec:method_teacher} the teacher, Cosmos3-Nano. The alternative teachers used in the ablations are described alongside their results in Sec.~\ref{sec:exp_ablation}.

\subsection{Student and Base Objective}
\label{sec:method_student}
The student is a QwenGR00T policy: a Qwen3-VL-family vision-language backbone \cite{qwen3vl} followed by a GR00T-style flow-matching action expert \cite{gr00t}. Given observations $\mathbf{o}_t$ (one or multiple camera views, a language instruction, and proprioceptive state), the backbone produces final-layer hidden states $h(\mathbf{o}_t)\in\mathbb{R}^{L\times D_s}$, where $L$ is the number of prefix tokens (the image tokens of every view, the instruction tokens, and the state token) and $D_s$ is the hidden dim; $h_i(\mathbf{o}_t)\in\mathbb{R}^{D_s}$ denotes the state at position $i$. The action expert predicts a chunk of $K$ future actions $\mathbf{a}_{t:t+K}$ by velocity regression along a flow path. Let $\epsilon\sim\mathcal{N}(\mathbf{0},\mathbf{I})$ be the noise, $\tau\in[0,1]$ be the flow time, and $\mathbf{a}^\tau = \tau\epsilon + (1-\tau)\,\mathbf{a}_{t:t+K}$ be the linear interpolant between data and noise, the expert regresses the path velocity $\frac{\mathrm{d}\mathbf{a}^\tau}{\mathrm{d}\tau} = \epsilon - \mathbf{a}_{t:t+K}$,
\begin{equation}
\mathcal{L}_{\mathrm{act}} =
\mathbb{E}_{\mathbf{o}_t,\tau,\epsilon}\big\|v_\theta(\mathbf{a}^\tau, \tau, h(\mathbf{o}_t))
- (\epsilon - \mathbf{a}_{t:t+K})\big\|^2 ,
\label{eq:act}
\end{equation}
where $\theta$ is the backbone and action-expert parameters.
Note that $\mathcal{L}_{\mathrm{act}}$ is supervised by \emph{ground-truth} demonstration actions throughout; no teacher action targets are used. We deliberately exclude the action channel: teacher and student do not share an action parameterization, and a recipe that depends on teacher actions cannot be teacher-agnostic. Everything the teacher contributes enters through its \emph{internal states}. The main student is a $0.8$B Qwen3.5-VL backbone; Sec.~\ref{sec:exp_ablation} repeats the recipe at $4$B and on a different backbone family. At inference the student runs one backbone prefill and four flow steps, and all distillation modules are dropped, so latency is identical to the undistilled student by construction.

\subsection{World-Model Representation Alignment}
\label{sec:method_c1}
Let $\mathcal{C}$ be the set of camera views. From $h(\mathbf{o}_t)$ we mean-pool the image-token span of each view $c\in\mathcal{C}$,
\begin{equation}
f^S_c(\mathbf{o}_t) = \frac{1}{|\mathcal{I}_c|}\sum_{i\in\mathcal{I}_c} h_i(\mathbf{o}_t)
\;\in\mathbb{R}^{D_s},
\end{equation}
where $\mathcal{I}_c\subseteq\{1,\dots,L\}$ collects the positions of view $c$'s image placeholder tokens within the prefix; the $\mathcal{I}_c$ are disjoint across views and together cover only the image portion of the $L$ positions. A two-layer MLP projector $p_\phi:\mathbb{R}^{D_s}\!\to\!\mathbb{R}^{D_t}$ maps this to the teacher width $D_t$ (Sec.~\ref{sec:method_teacher}), and we align it to the teacher's per-view pooled feature $f^T_c\in\mathbb{R}^{D_t}$, read from the cache, with a cosine loss,
\begin{equation}
\mathcal{L}_{\mathrm{align}} = 1-\cos\big(p_\phi(f^S_c(\mathbf{o}_t)),\ \mathrm{sg}[f^T_c(\mathbf{o}_t)]\big),
\label{eq:align}
\end{equation}
with $\cos(\mathbf{u},\mathbf{v})=\frac{\mathbf{u}^\top\mathbf{v}}{(\|\mathbf{u}\|\|\mathbf{v}\|)}$, $\mathrm{sg}[\cdot]$ the stop-gradient. Training minimizes Eq.~\eqref{eq:act} plus Eq.~\eqref{eq:align}, $\mathcal{L}_{\mathrm{act}} + \lambda_{\mathrm{align}} \mathcal{L}_{\mathrm{align}}$, with $\lambda_{\mathrm{align}}=0.5$ everywhere.

\textbf{Offline teacher cache.} The teacher is run once, ahead of training; targets are written to a memory-mapped cache keyed $(\text{trajectory id},\text{base index})$ order. Hence, no teacher weights are loaded during training, teacher and student can live in incompatible environments, and one cache serves multiple students, since the projector auto-sizes to whatever $(D_s,D_t)$ pair it is given. Each cached row is a single pooled vector per camera view, so the cache is cheap to store and cheap to produce: about an hour on four GPUs for LIBERO.

\subsection{Teacher: Cosmos3-Nano}
\label{sec:method_teacher}
Our teacher throughout is the understanding tower of Cosmos~3 \cite{cosmos3}, an omnimodal mixture-of-transformers world model. We extract its Qwen3-VL-8B reasoner from the unified checkpoint, dropping the other towers, and get the image-token hidden states at layer~24, mean-pooled per view, giving $D_t=4096$ per view. The tower is deterministic: there is no diffusion timestep or sampling. We also ablate two alternative teachers in Sec.~\ref{sec:exp_ablation}: Fast-WAM \cite{fastwam2026}, a world-action model built on the Wan2.2-TI2V-5B video diffusion transformer \cite{wan2025}, and V-JEPA2-AC \cite{assran2025vjepa2}, an action-conditioned predictive latent video model.

\section{Experiments}
\label{sec:exp}

We evaluate on two simulation benchmarks and two real robots, one single-arm and one bimanual. The question in each case is the same: does aligning a compact student to a frozen world model's representation buy accuracy that the same student cannot reach on its own, and how does the result compare to policies several times larger?

\subsection{Setup}
\label{sec:exp_setup}
\textbf{LIBERO} \cite{liu2023libero}: all four suites (\textsc{spatial}, \textsc{object}, \textsc{goal}, \textsc{libero-10}), the training data consists of $\sim$273k demonstration steps, two camera views. Checkpoints are evaluated on all four suites with $50$ episodes per task, i.e.\ $500$ trials per suite; we report the four-suite mean success rate $\mathrm{SR}_{\mathrm{avg}}$ in percent.

\textbf{RoboCasa-GR1} \cite{nasiriany2024robocasa,gr00t}: $24$ task environments of the GR1 Fourier humanoid (single ego camera, $29$-D bimanual action, $58$-D state), with $1{,}000$ demonstrations per environment, $24$k episodes and $6.02$M steps in total. We train one policy on all $24$ environments jointly and evaluate it on each of them, $20$ episodes per environment, reporting the mean of the per-environment success rates.

\textbf{Real robots}: an AgileX Nero run single-arm on two pick-and-place tasks, and TRIP-Bag \cite{tripbag2026}, a $7$-DoF bimanual platform, on a two-armed fruit handover (Sec.~\ref{sec:exp_real}).

\textbf{Training.} All students are trained on $4{\times}$A100-80GB with DeepSpeed ZeRO-2 at effective batch $256$, under a cosine schedule with minimum LR decaying over the run's full step budget.

\textbf{Evaluation protocol and its variance.} Neither simulation benchmark is deterministic. The policies sample actions from a flow-matching head, so two rollouts from one checkpoint differ, and the simulator's renderer is not bit-identical across GPU models, so the same checkpoint and the same seed can diverge on different hardware. Practitioners report both effects repeatedly \cite{nondet_libero,nondet_openvla,nondet_lerobot_a,nondet_lerobot_b,nondet_pulsevla,starvla_issue392}, but published tables are usually single numbers from a single machine. We therefore evaluate four times, crossing two evaluation seeds with two GPUs, an A100-80GB and an RTX~5090, and report the mean over those four runs with the standard deviation alongside. One interesting observation is that the two benchmarks are not equally stable. On LIBERO the spread is small, around a point to a point and a half, while on RoboCasa-GR1 the same checkpoint can move by several points when the GPU changes \cite{starvla_issue392}.

\subsection{LIBERO}
\label{sec:exp_libero}

Table~\ref{tab:libero} places our distilled $0.8$B student against published VLA and world-model policies. It reaches $97.9\%$ average success, $2.6$ points above the identical undistilled student. Our distilled model is still a touch behind the state-of-the-art VLA and World Model, however, the comparison at our own scale is the sharpest: every other policy below $4$B in this table sits between $78.7\%$ and $95.3\%$.

\begin{table}[!t]
\caption{LIBERO, per-suite success rate (\%). Baselines are quoted from their respective papers; our rows are the mean over four evaluation runs ($2$ seeds $\times$ $2$ GPUs).}
\label{tab:libero}
\centering
\scriptsize
\setlength{\tabcolsep}{1.2pt}
\begin{tabular}{@{}llrrrrr@{}}
\toprule
\textbf{Method} & \textbf{Size} & \textbf{Spat.} & \textbf{Obj.} &
\textbf{Goal} & \textbf{Long} & \textbf{Avg.} \\
\midrule
\multicolumn{7}{@{}l}{\textit{VLA policies}} \\
OpenVLA \cite{kim2025openvla}          & 7B   & 84.7 & 88.4 & 79.2 & 53.7 & 76.5 \\
Seer \cite{tian2025seer}               & 0.57B & -- & -- & -- & 78.7 & 78.7 \\
$\pi_0$-FAST \cite{pertsch2025fast}    & 3B   & 96.4 & 96.8 & 88.6 & 60.2 & 85.5 \\
SmolVLA \cite{shukor2025smolvla}       & 2.2B & 93.0 & 94.0 & 91.0 & 77.0 & 88.8 \\
GR00T~N1 \cite{gr00t}                  & 2B   & 94.4 & 97.6 & 93.0 & 90.6 & 93.9 \\
$\pi_0$ \cite{black2025pi0}            & 3B   & 96.8 & 98.8 & 95.8 & 85.2 & 94.2 \\
$\pi_{0.5}$-KI \cite{driess2026knowledge} & 3B & 98.0 & 97.8 & 95.6 & 85.8 & 94.3 \\
OpenVLA-OFT \cite{kim2025openvlaoft}   & 7B   & 97.6 & 98.4 & 97.9 & 94.5 & 97.1 \\
S$^2$-VLA \cite{s2vla}                 & 2B   & \textbf{98.4} & \textbf{99.6} & \textbf{98.4} & \textbf{96.4} & 98.2 \\
Abot-M0 \cite{abotm0}             & 4B & -- & -- & -- & -- & 98.6 \\
Qwen-RobotManip \cite{qwenrobotmanip}             & 4B & -- & -- & -- & -- & \textbf{99.2} \\
\midrule
\multicolumn{7}{@{}l}{\textit{World-model policies}} \\
WorldVLA \cite{cen2025worldvla}        & 7B   & 87.6 & 99.2 & 83.4 & 60.0 & 81.8 \\
Fast-WAM \cite{fastwam2026}             & 6B   & 98.2 & \textbf{100.0} & 97.0 & 95.2 & 97.6 \\
LingBot-VA \cite{li2026lingbot}        & 5.3B & 98.5 & 99.6 & 97.2 & \textbf{98.5} & 98.5 \\
Light-WAM \cite{lightwam}              & 2B   & 98.2 & 99.6 & \textbf{97.8} & 93.0 & 97.2 \\
Cosmos3-Nano \cite{cosmos3}             & 16B & -- & -- & -- & 97.6 & -- \\
Being-H0.7 \cite{beingh07}             & -- & -- & -- & -- & -- & \textbf{99.2} \\
\midrule
\multicolumn{7}{@{}l}{\textit{Ours}} \\
QwenGR00T, no distillation \cite{starvla2026} & 0.8B & 96.6\std{0.9} & 96.4\std{1.1} & 95.6\std{0.8} & 92.6\std{1.4} & 95.3\std{0.7} \\
\textbf{Ours}                          & 0.8B & \textbf{99.3}\std{0.6} & \textbf{99.2}\std{0.7} & \textbf{99.4}\std{0.5} & \textbf{93.8}\std{1.1} & \textbf{97.9}\std{0.5} \\
\bottomrule
\end{tabular}
\end{table}

\subsection{RoboCasa-GR1}
\label{sec:exp_gr1}


\begin{table}[!t]
\caption{RoboCasa-GR1, $24$ environments. $^\dagger$Re-evaluated by us from the released checkpoint under the same four-run protocol rather than quoted, because success rates shift across seeds and GPU devices \cite{starvla_issue392}.}
\label{tab:gr1}
\centering
\scriptsize
\setlength{\tabcolsep}{5pt}
\begin{tabular}{@{}lcc@{}}
\toprule
\textbf{Method} & \textbf{Size} & \textbf{SR (\%)} \\
\midrule
QwenFAST \cite{starvla2026}            & 4B   & 39.0 \\
QwenPI \cite{starvla2026}              & 4B   & 43.9 \\
Isaac-GR00T~N1.6 \cite{gr00t}          & 3B   & 47.6 \\
Isaac-GR00T~N1.5 \cite{gr00t}          & 3B   & 48.2 \\
QwenOFT \cite{starvla2026}             & 4B   & 48.8 \\
VP-VLA \cite{vpvla}                                 & 4B   & 53.8 \\
TwinBrainVLA \cite{twinbrainvla}       & 2$\times$4B & 54.6 \\
QwenGR00T$^\dagger$ \cite{starvla2026} & 4B   & 54.8\std{2.0} \\
ABot-M0 \cite{abotm0}                  & 4B   & 58.3 \\
PhysBrain \cite{physbrain}             & 4B   & 64.5 \\
ACE-Ego-0 \cite{aceego0}               & 4B   & 72.8 \\
\midrule
QwenGR00T, no distillation \cite{starvla2026} & 0.8B & 48.2\std{2.1} \\
\textbf{Ours} & 0.8B & \textbf{50.5}\std{2.3} \\
\bottomrule
\end{tabular}
\end{table}

Table~\ref{tab:gr1} isolates the contribution of the method: with the alignment term switched on, the same $0.8$B student improves from $48.2\%$ to $50.5\%$ success, a gain of $2.3$ points, at zero inference cost. That is enough to move a $0.8$B policy past QwenFAST, QwenPI and QwenOFT and both Isaac-GR00T releases, and to within $4.3$ points of the $4$B QwenGR00T ($54.8\%$), of which our student is a five-fold reduction and which the same objective lifts again when applied to that backbone directly (Table~\ref{tab:students}); the stronger published entries, all at $4$B or more, remain ahead.

The stronger entries get there by building something: ACE-Ego-0 assembles roughly $600$M frames of robot and pseudo-action-labelled human video through a five-stage pipeline \cite{aceego0}, and PhysBrain a question-answer corpus, a retrained base VLM, and a dual-pathway adaptation architecture \cite{physbrain}. We rely on large-scale pretraining too, but only through a world model that already exists and that we neither train nor deploy: the recipe adds one additional simple loss, needs no data beyond the original training data, and leaves the deployed policy unchanged.

\subsection{Real-Robot Manipulation}
\label{sec:exp_real}

Simulation results are only as good as their transfer, so we validate on hardware. The evaluation is built to vary one thing at a time. We begin on an AgileX Nero, a $7$-DoF arm operated single-arm, with two separate pick-and-place tasks, one on fruit and one on eggs; we then carry the fruit task over to TRIP-Bag \cite{tripbag2026}, a $7$-DoF bimanual platform, and rebuild it as a two-armed handover. The second step changes the embodiment, the number of arms, the cameras, and the controller latencies at once, so it asks whether the gain is a property of the method or of one particular setup.

\textbf{Task 1: fruit pick-and-place (single-arm).} Fruits lie on the table in front of the robot and a basket sits at the edge of the workspace. The arm must locate each fruit in turn, grasp it, lift it clear of the table, and release it into the basket.

\textbf{Task 2: egg pick-and-place (single-arm).} The two eggs sit in a crate and must be moved one at a time into a basket. An egg is smooth, close to spherical, and slippery, so a grasp that would hold a fruit slides off it, and the shell tolerates only a narrow band of closing force. The difficulty is in how the object responds to contact rather than in finding it.

\textbf{Task 3: bimanual fruit handover (TRIP-Bag).} Three fruits lie on the table and an open bag sits at the left edge of the workspace. For each fruit, the \emph{right} arm must locate and grasp it, lift it clear of the table, hand it over to the \emph{left} gripper in mid-air, and the left arm must then release it into the bag. The policy repeats this for all three fruits. Making the handover explicit forces the two arms into a timed dependency, and the bag is deformable, so its opening geometry changes as it fills.

In all three tasks the objects are cleared one at a time and a trial counts as a success only if every one of them ends up inside the target receptacle, so an object dropped on the way or left resting on the rim fails the trial outright. We fine-tune each policy per platform on roughly $30$~min of teleoperated demonstration for the single-arm tasks and $1$~h for the bimanual one, and score $30$ trials per policy, task, and platform.

\begin{figure*}[!t]
    \centering
    \includegraphics[width=\textwidth]{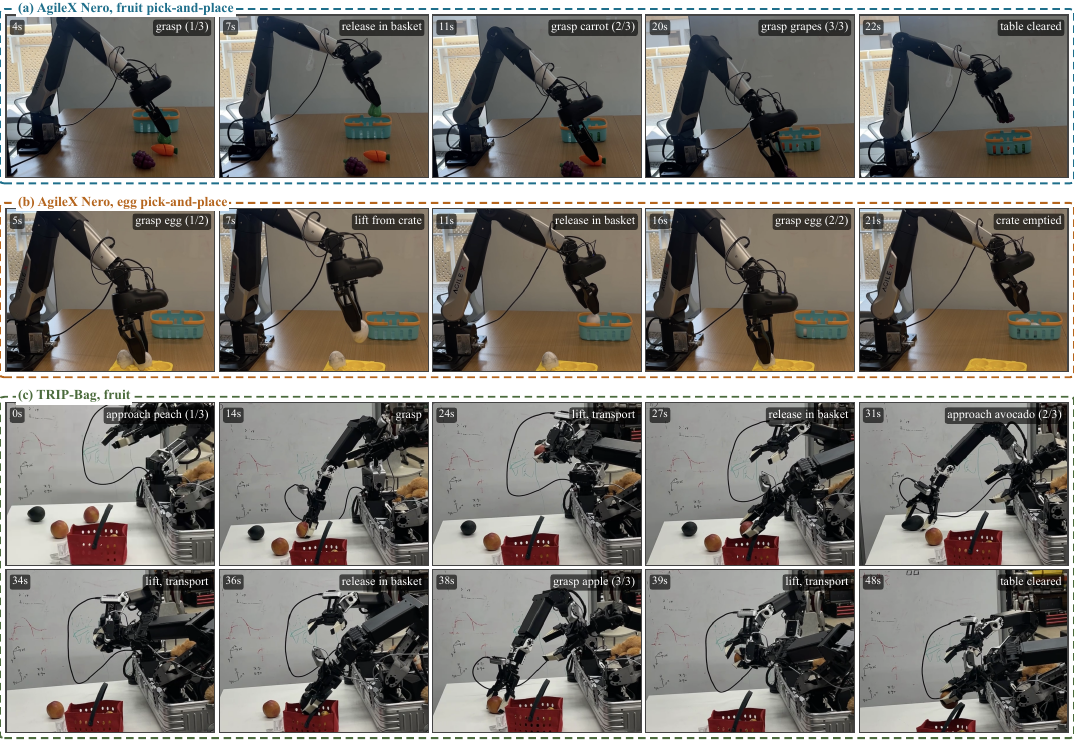}
    \caption{Successful rollouts of the distilled $0.8$B student on the three real-robot tasks, recorded from a third-person camera. Each dashed box is one episode: \textbf{(a)}~fruit and \textbf{(b)}~egg pick-and-place on the AgileX Nero, and \textbf{(c)}~the fruit task on TRIP-Bag.}
    \label{fig:real_rollout}
\end{figure*}

\begin{table}[!t]
\caption{Real-robot evaluation, success rate (\%) over $30$ trials per cell.}
\label{tab:real}
\centering
\scriptsize
\setlength{\tabcolsep}{4pt}
\begin{tabular}{@{}lcccc@{}}
\toprule
 & & \multicolumn{2}{c}{\textbf{AgileX Nero} (1-arm)} & \textbf{TRIP-Bag} (2-arm) \\
\cmidrule(lr){3-4}\cmidrule(lr){5-5}
\textbf{Policy} & \textbf{Params} & \textbf{Fruit} & \textbf{Egg} & \textbf{Fruit} \\
\midrule
$\pi$-style policy      & 4B   & \textbf{93.3} & \textbf{70.0} & \textbf{53.3} \\
\midrule
QwenGR00T, no distillation & \textbf{0.8B} & 83.3 & 46.7 & 40.0 \\
\textbf{Ours}           & \textbf{0.8B} & \textbf{93.3} & 60.0 & 46.7 \\
\bottomrule
\end{tabular}
\\[2pt]
\end{table}

Fig.~\ref{fig:real_rollout} shows a successful rollout, and Table~\ref{tab:real} the scores. The same pattern as GR1 holds, and the three columns order themselves by difficulty. On the single-arm fruit task the distilled $0.8$B policy reaches $93.3\%$ ($28/30$), matching the $4$B $\pi$-style policy and $3$ trials above the undistilled control. On the egg task every policy drops, mostly due to slippage during grasping the eggs. The gap that opens is the informative part: the control falls to $46.7\%$ ($14/30$) while distillation recovers $4$ of the $7$ trials separating it from the $4$B policy, reaching $60.0\%$ ($18/30$) against $70.0\%$ ($21/30$). Moving to two arms costs every policy again, and the same relation holds, $40.0\%$ for the control, $46.7\%$ for ours, $53.3\%$ for the $4$B policy.

\begin{figure}[!t]
    \centering
    \includegraphics[width=\columnwidth]{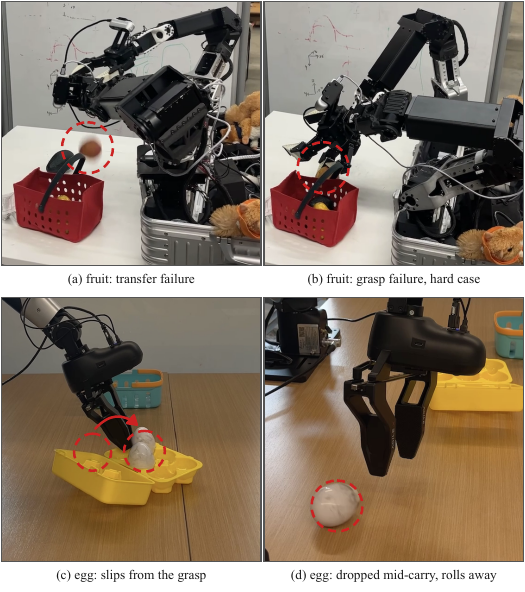}
    \caption{Failure modes, fruit (top) and egg (bottom); the ring marks the failure. \textbf{(a)}~Dropped before the receptacle. \textbf{(b)}~Closed on the rim, not the fruit. \textbf{(c)}~Egg slips from the grasp. \textbf{(d)}~Egg slips mid-carry.}
    \label{fig:real_failures}
\end{figure}

\textbf{Failure modes.} The failures are not uniform, and the modes that dominate are shown in Fig.~\ref{fig:real_failures}. All are errors of \emph{placement} rather than of recognition: the policy consistently finds the target and reaches for the right one, and what it misses is the last few centimetres, either letting go early on the way to the receptacle or closing on the receptacle's own geometry when the two are close enough to overlap in the image. The egg task adds a grasp variant of the same error: the fingers close on the egg and lift, but the shell slips sideways (Fig.~\ref{fig:real_failures}c). When an egg slips out mid-carry (Fig.~\ref{fig:real_failures}d), the policy still tracks it and reaches for it again, so it is the contact that fails, not the scene understanding. This is what we would expect a representational prior to help with least, since none of these modes is a failure to understand the scene.

\subsection{Ablations}
\label{sec:exp_ablation}

The ablations ask which parts of the recipe matter. The student and alignment-layer ablations are run on RoboCasa-GR1, which has the headroom to separate the variants; the teacher ablation is run on LIBERO to utilize existing pretrained teacher models. The RoboCasa-GR1 ablations are scored only once on A100 due to limited computational resources.

\textbf{Student scale and backbone} (Table~\ref{tab:students}). The recipe is defined on the student's pooled image-token hidden states and nothing else, so it carries across student sizes and backbone families without modification: the projector auto-sizes to whatever student width it is given. We therefore repeat it at $4$B within the same backbone family and at $1$B on a different one.

\begin{table}[!t]
\caption{Student scale and backbone family ablation on RoboCasa-GR1. \textbf{w/o} is the same student trained without the alignment term. Single A100 run, so these differ from the four-run means of Table~\ref{tab:gr1}.}
\label{tab:students}
\centering
\footnotesize
\setlength{\tabcolsep}{5pt}
\begin{tabular}{@{}llcc@{}}
\toprule
\textbf{Student backbone} & \textbf{Params} & \textbf{w/o} & \textbf{SR} (\%) \\
\midrule
Qwen3.5-VL   & 0.8B & 50.3 & \textbf{52.8} \\
InternVL     & 1B   & 50.9 & 53.3 \\
Qwen3-VL     & 4B   & 56.8 & 58.4 \\
\bottomrule
\end{tabular}
\end{table}

\textbf{Which student layer to align} (Table~\ref{tab:taplayer}). The alignment term needs one hidden state from the student, and nothing in the derivation says which. We sweep the alignment point across the depth of the $0.8$B student, at a third, a half, two thirds, and the final layer. To keep the sweep affordable we run each variant on RoboCasa-GR1 for a quarter of the full schedule, so the absolute numbers sit below Table~\ref{tab:gr1} and only the ordering is meaningful. Every choice trains stably and lands within four points of the best, but the ordering is not monotonic in depth: the final layer is best at $49.8\%$, the half-depth layer is within half a point of it at $49.4\%$, and the two-thirds layer is the weakest at $45.9\%$. We read this as the objective being tolerant of where it is attached rather than as evidence for a particular depth, since the spread between the best and second-best layers is smaller than what a quarter-length schedule can resolve. We keep the final layer, which is also the cheapest to implement: it is the representation the action head already consumes, so no intermediate layer's activation has to be retained for loss computation.

\begin{table}[!t]
\caption{Student layer ablation on RoboCasa-GR1. Trained for a quarter of the full schedule and evaluated once on an A100, so the rates sit below Table~\ref{tab:gr1}.}
\label{tab:taplayer}
\centering
\footnotesize
\setlength{\tabcolsep}{5pt}
\begin{tabular}{@{}llc@{}}
\toprule
\textbf{Aligned layer} & \textbf{Depth} & \textbf{SR} (\%) \\
\midrule
L8            & $1/3$  & 47.5 \\
L12           & $1/2$  & 49.4 \\
L16           & $2/3$  & 45.9 \\
L24 (final)   & full   & \textbf{49.8} \\
\bottomrule
\end{tabular}
\end{table}

\textbf{Teacher} (Table~\ref{tab:teachers}). The claim we want to test is that \emph{a} world model helps, not that \emph{this} world model helps, so we repeat the recipe with two teachers chosen to differ from Cosmos3-Nano, and from each other, on every axis available: training objective, architecture, and feature width. \textbf{Fast-WAM} \cite{fastwam2026} is a world-action model built on the Wan2.2-TI2V-5B video diffusion transformer \cite{wan2025}; we read the frame-0 hidden state of its video expert at layer~15 of~30. \textbf{V-JEPA2-AC} \cite{assran2025vjepa2} pairs a self-supervised ViT-g/16 video encoder with an action-conditioned predictor; we post-train it on the target dataset and read the predictor's pre-projection hidden state. 

Every teacher improves on the undistilled control of Table~\ref{tab:libero} ($95.3$): V-JEPA2-AC by $1.2$ points, Fast-WAM by $1.6$, and Cosmos3-Nano by $2.6$. What matters most is the consistency across teachers: since all three lift the same student, the transferred signal is best attributed to world-model representations in general rather than to Cosmos3-Nano specifically. The benefit is thus a property of the recipe rather than of a particular teacher.


\begin{table}[!t]
\caption{Teacher ablation on LIBERO.}
\label{tab:teachers}
\centering
\scriptsize
\setlength{\tabcolsep}{3pt}
\begin{tabular}{@{}lccc@{}}
\toprule
 & \textbf{Cosmos3-Nano} & \textbf{Fast-WAM} & \textbf{V-JEPA2-AC} \\
 & \textbf{(main)} & \textit{(ablation)} & \textit{(ablation)} \\
\midrule
family & omnimodal VLM & video DiT & enc.--predictor \\
base & Cosmos~3 MoT & Wan2.2-TI2V-5B & ViT-g/16 + AC \\
size & 8B tower (of 16B) & 5B & 1.0B + 0.3B \\
read at & LM layer 24 & DiT layer 15/30 & predictor norm \\
$D_t$ (per view) & 4096 & 3072 & 1024 \\
\midrule
\multicolumn{4}{@{}l}{\textit{Distilled $0.8$B student, LIBERO (no distillation: $95.3$)}} \\
$\mathrm{SR}_{\mathrm{avg}}$ (\%) & \textbf{97.9}\std{0.5} & 96.9\std{0.8} & 96.5\std{0.3} \\
\bottomrule
\end{tabular}
\end{table}


\section{Conclusion}
\label{sec:conclusion}

World models earn their grounding by predicting the future, and that same prediction is what keeps them out of a control loop. We showed the two can be pulled apart. A compact policy aligned once, offline, to a frozen world model's visual features keeps some of what that grounding is worth while deploying exactly the network it would have deployed anyway, at no cost in latency or memory. The gains are modest but consistent, in both simulation environments (LIBERO, RoboCasa-GR1) and real-robot experiments, and they hold across student scales, backbones, alignment layers, and teachers, which is the pattern one would expect from a representational prior rather than from a coincidence between two networks.

\section*{Acknowledgment}

This work was supported in part by NSF IIS2404180, National Research Foundation of Korea (NRF) grant (MSIT) (RS-2026-25522067) and Institute of Information \& communications Technology Planning\& Evaluation (IITP) grants funded by the Korea government (MSIT) (No. 2022-0-00871, Development of AI Autonomy and Knowledge Enhancement for AI Agent Collaboration), (No. RS-2025-2543949. Environment-Aware and Domain-Adaptive Multimodal Embodied AI for Real-World Interaction), and Electronics and Telecommunications Research Institute (ETRI) grant (26CB1200, Development and Application of Science-Specialized Multimodal Foundation Models). This work also used Delta and DeltaAI at the University of Illinois Urbana-Champaign through allocation CIS261053 from the Advanced Cyberinfrastructure Coordination Ecosystem: Services \& Support (ACCESS) program, which is supported by U.S. National Science Foundation grants \#2138259, \#2138286, \#2138307, \#2137603, and \#2138296. This work also used computing resources made available through the AMD University Program (AUP) AI \& HPC Cluster.

\bibliographystyle{IEEEtran}
\bibliography{main}

\end{document}